# Case Study: Testing Model Capabilities in Some Reasoning Tasks

Min Zhang, Sato Takumi, Jack Zhang, Jun Wang

**Abstract**

Large Language Models (LLMs) excel in generating personalized content and facilitating interactive dialogues, showcasing their remarkable aptitude for a myriad of applications. However, their capabilities in reasoning and providing explainable outputs, especially within the context of reasoning abilities, remain areas for improvement. In this study, we delve into the reasoning abilities of LLMs, highlighting the current challenges and limitations that hinder their effectiveness in complex reasoning scenarios.

## 1 Introduction

In the rapidly evolving field of artificial intelligence [30], Large Language Models (LLMs) have emerged as a cornerstone of technological advancement, revolutionizing the way we interact with machines and process information. With their unparalleled ability to generate human-like text, LLMs have found applications across a broad spectrum of domains, from automating customer service interactions to aiding in the creative process of writing and design. Their proficiency in generating personalized content and facilitating interactive dialogues has underscored their versatility and adaptability, making them indispensable tools in the modern digital landscape [4, 5, 6].

Despite these significant achievements, LLMs are not without their shortcomings. One of the critical areas where LLMs still face challenges is in their reasoning abilities and the provision of explainable outputs. While these models excel at tasks that require understanding and generating natural language, their performance in scenarios that demand complex reasoning and the ability to articulate the underlying logic of their conclusions is less robust [18, 1]. This limitation not only impacts the reliability of LLMs in decision-making processes but also raises concerns about their transparency and the trustworthiness of their outputs.

The ability to reason, a fundamental aspect of intelligence, involves more than just processing information; it encompasses the capability to draw inferences, make connections between disparate pieces of data, and apply knowledge to novel situations. For LLMs to reach their full potential and become more integrated into critical decision-making processes, enhancing their reasoning capabilities is paramount. However, achieving this improvement is not straightforward, as it requires advancements in both the models' architecture and the methodologies used for their training and fine-tuning [15].

In this study, we delve into the intricacies of the reasoning abilities of LLMs, exploring the current challenges and limitations that undermine their effectiveness in complex reasoning scenarios. Our investigation reveals that while LLMs have made significant strides in various aspects of natural language understanding and generation, their performance in tasks requiring advanced reasoning, particularly those that necessitate an understanding of causality, logic, and complex problem-solving, remains an area ripe for enhancement [2].

To address these challenges, we propose a multifaceted approach that aims to bolster the reasoning capabilities of LLMs. Our strategy emphasizes the importance of parameter-efficient fine-tuning methods and advanced prompting strategies, which together can significantly improve the models' ability to engage in sophisticated reasoning tasks. Central to our approach is the introduction of a novel model, ReasonAlpaca, which epitomizes the integration of these methodologies. ReasonAlpaca is fine-tuned using the Low-rank Adaptation (LoRA) technique on LMs, a model already renowned for its proficiency in language tasks. By utilizing a specialized 13k instruction-following dataset for fine-tuning, we tailor the model's capabilities towards enhanced reasoning performance [9].

The results of our study are both promising and illuminating. ReasonAlpaca demonstrates a marked improvement in reasoning accuracy over existing baselines, showcasing the effectiveness of our proposed approach. Through a series of rigorous evaluations, we establish that the integration of LoRA and targeted fine-tuning using instruction-following datasets can significantly enhance the reasoning abilities of LLMs. Our findings not only contribute to the ongoing discourse on improving LLMs but also pave the way for future research endeavours aimed at unlocking the full potential of these models in complex reasoning scenarios [29].

## 2 Method and Baselines

State-of-the-art Parameter-Efficient Fine-Tuning (PEFT) methods have demonstrated remarkable capabilities, achieving performance levels on par with those of full model fine-tuning while updating only a fraction of the model's parameters. This approach marks a significant advancement in model efficiency, facilitating rapid adaptation to specific tasks without necessitating extensive retraining of the entire model architecture. Despite these advancements, PEFT methodologies encounter challenges when applied alongside continued pre-training processes [8, 13, 7, 21]. These challenges limit their flexibility in further tailoring Language Models (LMs) for more specialized or nuanced tasks, which could benefit from additional, focused pre-training [11, 17, 24, 14].

Within the domain of PEFT, Prompt Tuning has emerged as an innovative strategy, designed to enhance the model's adaptability to specific tasks through the introduction of task-specific tokens [16]. As detailed by Shi et al. [23], this method capitalizes on the model's existing architecture by directly embedding these specialized tokens into the input layer. This technique facilitates direct training with task-specific supervision, improving the model's performance on targeted tasks without requiring significant updates to its parameters. However, the success of Prompt Tuning heavily depends on the meticulous initialization of these prompt tokens. [11] have identified the challenges linked to random initialization, noting its potential to lead to suboptimal results. To overcome this, they recommend initializing the embeddings of these tokens based on pre-selected prompt tokens, identified through empirical or heuristic methods to ensure task-specific relevance [11, 17, 27, 31].

This discussion highlights the delicate equilibrium between efficiency and customization in the application of PEFT methods such as Prompt Tuning. Although these methods provide a pathway toward more resource-efficient model adaptation, the complexities involved in token initialization and the overall compatibility with ongoing pre-training frameworks open up opportunities for further research and innovation. Addressing these challenges could pave the way for future advancements in PEFT, unlocking even greater possibilities for customizing LMs to a broader range of specific tasks, thus expanding the horizons of natural language processing capabilities. Moreover, the integration of data augmentation strategies, as proposed by Wei et al. [31] and further elaborated by [22, 12], alongside techniques from Sun et al. [27] and innovations in adaptive learning rates and model efficiency [7, 14], contribute significantly to the refinement of PEFT methods. These enhancements, by diversifying the training data and fostering model generalization across various contexts, are instrumental in overcoming the limitations of current PEFT approaches, driving the field toward achieving nuanced and highly effective language models.

## 3  Problem Statement.

The challenge at hand involves a task structure that is parallel to established benchmarks within the field, specifically focusing on tasks that necessitate a foundation in commonsense knowledge. In essence, the task is formulated around a given natural language question or description, denoted as $q$. Accompanying $q$ is a collection of $N$ potential answers, represented as $a_i$. The primary objective is to accurately identify the most appropriate answer from this set. In certain scenarios, an additional context $c$ is provided, serving as a supplementary aid in the decision-making process.

Furthermore, the task is augmented by the presence of a large external knowledge graph, $G$. This knowledge graph plays a crucial role, offering a rich source of information that can be leveraged to assess the viability of the answer candidates. It contains interconnected data that can illuminate the relationships and facts relevant to the question or description at hand, thus enabling a more informed selection process.

The overarching goal of this task is to accurately determine the correct answer from the set $a_i$ by effectively synthesizing the information presented in $q$, the optional context $c$, and the insights derived from the external knowledge graph $G$. This requires not only a deep understanding of the language and the nuances of the question but also an ability to navigate and utilize the complex, structured information within $G$. The task underscores the importance of integrating diverse sources of information, including direct question-answer matching, contextual interpretation, and external knowledge validation, to achieve a comprehensive understanding and accurate decision-making in scenarios where commonsense reasoning is paramount.

To address this challenge, it is imperative to develop methodologies that can seamlessly integrate natural language processing techniques with knowledge graph analytics. Such approaches must be capable of discerning the subtle semantics of the question and context, extracting and correlating relevant information from the knowledge graph, and applying commonsense reasoning to evaluate the plausibility of each answer candidate. This multifaceted task highlights the intersection of language understanding, knowledge representation, and reasoning, pushing the boundaries of current capabilities in artificial intelligence and natural language processing.

## 4 Tasks

To evaluate intelligent agents' commonsense reasoning ability, many benchmarks have been proposed. Here we summarise several most widely-used benchmarks, with examples of each benchmark.

paragraphOverview of Commonsense Reasoning Benchmarks The evaluation of intelligent agents' ability to perform commonsense reasoning is a critical area of research in artificial intelligence. To this end, a variety of benchmarks have been established, each designed to test different facets of commonsense understanding in natural language processing (NLP). These benchmarks, which are primarily structured as multiple-choice questions, require agents to not only understand the given information but also to draw upon external commonsense knowledge to arrive at the correct answer. Below, we provide a summary of several prominent benchmarks in this domain.

- **MCTACO** [32] focuses on temporal commonsense, containing questions about the timing and sequence of events. This dataset introduces a unique dimension to commonsense reasoning tasks.

- **PIQA** [3] targets physical commonsense understanding, providing over 16,000 question-answer pairs related to everyday physical phenomena. The human benchmark for this dataset is 94.9

- Building on CommonsenseQA, Rajani et al. [19] introduced **commonsense Explanations (CoS-E)**, adding human explanations to the questions and answers to facilitate understanding of the reasoning process.

- **StepGame** [25] is designed to test spatial reasoning capabilities within NLP models, challenging agents to understand and infer spatial relationships based on natural language descriptions.

- **Cosmos QA** [10] features 35,588 questions that demand commonsense-based reading comprehension, focusing on understanding the causes and effects within narratives. This dataset explores a wide range of inferential reasoning, with a human performance level of 94

- **Winogrande** [20] extends the Winograd Schema Challenge by including a larger set of problems to combat dataset-specific biases and test commonsense reasoning more rigorously. Human performance here is 94

- **CommonsenseQA** [28] presents a challenge with 12,247 questions that necessitate commonsense knowledge, specifically leveraging the ConceptNet knowledge graph [26]. The dataset tests an agent's ability to utilize broad commonsense knowledge, achieving a human benchmark performance of 94.1

These benchmarks not only test an agent's ability to understand and process natural language but also require the integration of external knowledge—such as facts from ConceptNet or specific domain knowledge—to correctly answer the questions. In the next section, we delve into the methodologies for representing and leveraging this external knowledge to enhance commonsense reasoning capabilities in NLP models.

# 5 Discussion

The exploration and evaluation of intelligent agents' commonsense reasoning capabilities through a diverse array of benchmarks have highlighted the significant strides made in the field of artificial intelligence, specifically within natural language processing. The benchmarks discussed, ranging from CommonsenseQA [28] to StepGame [25], each present unique challenges that probe the depth and breadth of an agent's understanding of commonsense knowledge and its application in various contexts. These tasks underscore the necessity for models to not only parse and understand natural language but also to draw upon a vast reservoir of external knowledge and reasoning abilities to navigate complex, real-world scenarios effectively.

The integration of external knowledge, exemplified by resources such as ConceptNet [26], and the application of advanced NLP techniques have been central to advancing the performance of intelligent agents on these benchmarks. However, the disparities in performance between human benchmarks and current models illuminate the ongoing challenges in achieving truly robust and nuanced commonsense reasoning capabilities in machines.

Future research directions are poised to address these challenges through more sophisticated models, innovative training methodologies, and the integration of even richer sources of external knowledge. The ultimate goal remains to bridge the gap between human-like understanding and machine processing, enabling intelligent agents to perform with a level of subtlety, flexibility, and depth that mirrors human reasoning. As the field progresses, the continuous refinement of benchmarks and the development of models that can navigate the intricacies of commonsense reasoning will be pivotal in pushing the boundaries of what artificial intelligence can achieve.

In this journey, the collaboration between diverse domains within artificial intelligence, from knowledge representation and reasoning to machine learning and natural language processing, will be crucial. By fostering interdisciplinary approaches and leveraging the collective insights from these fields, the development of intelligent agents with advanced commonsense reasoning capabilities will not only enhance our understanding of artificial intelligence but also expand its potential applications, making technology more intuitive, responsive, and beneficial to society.